\documentclass[letterpaper]{article} 
\usepackage{aaai24}  
\usepackage{times}  
\usepackage{helvet}  
\usepackage{courier}  
\usepackage[hyphens]{url}  
\usepackage{graphicx} 
\usepackage{natbib}  
\usepackage{caption} 
\usepackage{algorithm}
\usepackage{algorithmic}
\usepackage{amsfonts,amssymb}
\usepackage{tabularx}
\usepackage{booktabs}
\usepackage{newfloat}
\usepackage{listings}
\usepackage{amsmath}
\usepackage{longtable}
\usepackage{multirow}
\DeclareCaptionStyle{ruled}{labelfont=normalfont,labelsep=colon,strut=off} 
\floatstyle{ruled}
\newfloat{listing}{tb}{lst}{}
\floatname{listing}{Listing}
\title{A Progressive Framework of Vision-language Knowledge Distillation and Alignment for Multilingual Scene}

\author{
    Wenbo Zhang \textsuperscript{\rm a} \textsuperscript{\rm 1},
    Yifan Zhang \textsuperscript{\rm a} \textsuperscript{\rm 1},
    Jianfeng Lin \textsuperscript{\rm b},
    Binqiang Huang \textsuperscript{\rm b},\\
    Jinlu Zhang \textsuperscript{\rm b},
    Wenhao Yu \textsuperscript{\rm a} \textsuperscript{\rm c} \textsuperscript{\rm *} 
}
\affiliations{
    \textsuperscript{\rm a}School of Geography and Information Engineering, China University of Geosciences, Wuhan, China\\
    \textsuperscript{\rm b}Meituan, Beijing, China\\
    \textsuperscript{\rm c}National Engineering Research Center for Geographic Information System, China University of Geosciences, Wuhan, China\\
    \textsuperscript{\rm 1}These authors contributed equally\\
    \textsuperscript{\rm *}Corresponding author, email: yuwh@cug.edu.cn\\

}

\usepackage{bibentry}

\begin{document}

\maketitle

\begin{abstract}
Pre-trained vision-language (V-L) models such as CLIP have shown excellent performance in many downstream cross-modal tasks. However, most of them are only applicable to the English context. Subsequent research has focused on this problem and proposed improved models, such as CN-CLIP and AltCLIP, to facilitate their applicability to Chinese and even other languages. Nevertheless, these models suffer from high latency and a large memory footprint in inference, which limits their further deployment on resource-constrained edge devices. In this work, we propose a conceptually simple yet effective multilingual CLIP Compression framework and train a lightweight multilingual vision-language model, called DC-CLIP, for both Chinese and English context. In this framework, we collect high-quality Chinese and English text-image pairs and design two training stages, including multilingual vision-language feature distillation and alignment. During the first stage, lightweight image/text student models are designed to learn robust visual/multilingual textual feature representation ability from corresponding teacher models, respectively. Subsequently, the multilingual vision-language alignment stage enables effective alignment of visual and multilingual textual features to further improve the model's multilingual performance. Comprehensive experiments in zero-shot image classification, conducted based on the ELEVATER benchmark, showcase that DC-CLIP achieves superior performance in the English context and competitive performance in the Chinese context, even with less training data, when compared to existing models of similar parameter magnitude. The evaluation demonstrates the effectiveness of our designed training mechanism.
\end{abstract}

\section{Introduction}
Learning a good representation in the combined domain of vision and language remains a pivotal goal of foundational model research. The recent release of CLIP \cite{[1]} by OpenAI is a landmark work in the field of visual language. Trained on extensive image-text pairing data, CLIP learns to map images and arbitrary textual descriptions into a shared feature space, thereby enabling the model to comprehend and process a variety of visual tasks, including image classification \cite{[2],[3],[4]}, object detection \cite{[5],[6],[7]}, and semantic segmentation \cite{[8],[9]}, without necessitating task-specific training data. This pre-training methodology significantly enhances the model's generalization capabilities and adaptability, facilitating its application across a broad spectrum of visual understanding tasks and showcasing its remarkable zero-shot transfer capability across various downstream tasks.
\begin{figure*}
  \centering
 \includegraphics[width=0.75\linewidth]{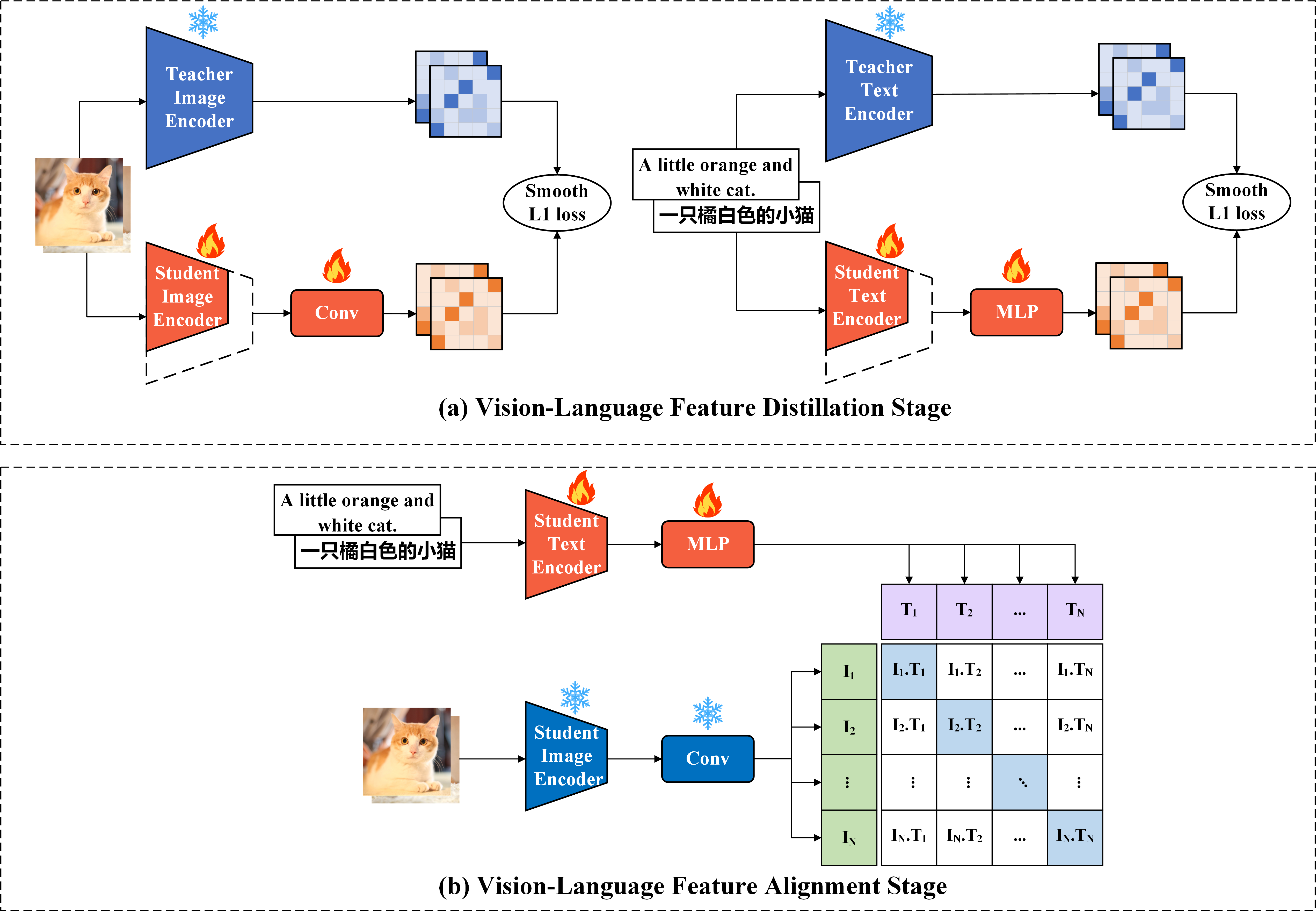}
 \caption{The framework of the proposed method comprises two progressive modules: (a) the vision-language feature distillation module and (b) the vision-language feature alignment module. In the first stage, module (a) extracts key knowledge from the teacher model's image and text encoders and transfers it to the student model. Subsequently, module (b) further aligns the vision-language feature using  contrastive learning strategy to enhance the model's performance.}
 \label{figure1}
\end{figure*}

However, CLIP's reliance on English text data for training introduces performance limitations in non-English linguistic contexts, prompting research into adapting CLIP for diverse language environments. For instance, Chinese CLIP \cite{[10]} leverages 123 million Chinese image-text pairs to retrain the model, improving the understanding of Chinese visual content. Multilingual CLIP \cite{[11]}, meanwhile, extends CLIP's applicability to multilingual settings by training multilingual text encoders with machine translation techniques and cross-linguistic teacher-learning methods. AltCLIP \cite{[12]} employs teacher learning and contrastive learning strategies to learn bilingual and multilingual multimodal representation models.

Despite the considerable strides made with the CLIP model and its adaptation under multilingual environment \cite{[10],[11],[12],[13]}, research on its effective deployment to mobile or edge devices remains scant \cite{[14],[15],[16]}. This gap primarily arises from the intrinsic constraints of such devices, including limited computational capacity, constrained storage, and stringent energy requirements, which pose formidable challenges to the design and deployment of efficient deep learning models, particularly complex multimodal ones like CLIP. In response, there is a pressing need for novel methodologies and technologies to optimize model size and computational demands, ensuring their operability on resource-constrained devices without compromising performance.

As one of the effective techniques for compressing large models, knowledge distillation \cite{[22],[48]} involves injecting knowledge from a powerful Teacher model into a smaller Student model without sacrificing much of its generalization ability. It has been extensively researched and applied in the single-modal environment \cite{[49]}. However, its potential in the multimodal domain remains largely untapped. Unlike single-modal models, distilling multimodal models, such as language-image cross-modal models like CLIP, presents distinct challenges. Firstly, these multimodal models, similar to CLIP, typically consist of two branches: an image encoder and a text encoder \cite{[50],[51],[52],[53]}. When distilling such multi-branch models, it is crucial to consider the information interaction between different modal branches in both the Teacher and Student models. Secondly, the original CLIP model was pre-trained on 400 million image-text pairs for 32 epochs, requiring thousands of GPU-days, which poses a significant challenge for distillation when computational resources are limited.

Addressing this challenge, our study introduces DC-CLIP, a novel, lightweight variant of the CLIP model devised through feature distillation \cite{[17]} and contrastive learning \cite{[18]}. Our approach is geared towards significantly reducing the model's demand for computational resources and storage space without sacrificing the model's performance, and at the same time realizing the application of CLIP in both the Chinese and English contexts. Specifically, the DC-CLIP consists of two main stages: the vision-language feature distillation stage and the vision-language alignment stage. Initially, we take the Chinese and English versions of AltCLIP as teacher models and distill their image encoders and text encoders respectively, which enables us to extract the key knowledge from the original AltCLIP model with a large number of parameters and transfer it effectively to a smaller and more efficient model architecture. Subsequently, the model undergoes refinement through contrastive learning to enhance its precision in image-text alignment within a multilingual framework. This phase predominantly relies on a modest corpus of Chinese and English image-text pairs, optimizing the model via contrastive loss to facilitate image-text matching and comprehension in multilingual contexts.

Through this methodology, DC-CLIP maintains robust performance in Chinese and English settings while adapting to the resource limitations of mobile devices such as smartphones. This breakthrough not only epitomizes technological innovation in multilingual vision-language models  but also carves new avenues for deploying efficient and pragmatic multimodal model applications on a plethora of mobile and edge devices in the future. With DC-CLIP, we anticipate offering users a more seamless and effective visual language interaction experience across multilingual landscapes, thereby broadening the applicability and adoption of the CLIP model and its derivatives on mobile platforms.

Our contributions are as follows:
\begin{itemize}
    \item Adaptability to English and Chinese: Innovatively adapts models for both English and Chinese contexts, enhancing linguistic inclusivity and comprehension across these languages.
    \item Feature Distillation: Employs output feature maps from pre-trained models as distillation targets, offering a richer knowledge transfer than traditional logits or simple vectors.
    \item Efficiency with Limited Data: We organized a small-scale, high-quality dataset of Chinese and English image-text pairs. Following training, our model demonstrates competitive performance in the Chinese context compared to the baselines and achieves state-of-the-art performance in the English context. This showcases the effectiveness and practicality of our approach in scenarios with limited data availability.
\end{itemize}

\section{Related work}
\begin{table*}[ht]
  \caption{Description of datasets}
  \begin{tabularx}{\textwidth}{lXXX} 
      \toprule
      Datasets      & Usage                                   & Component \\
      \toprule
      ImageNet      & Generic-objects datasets                & Humans, animals, vehicles, food, etc. \\
      Caltech101    & Generic-objects datasets                & Animals, plants, transportation, furniture, etc. \\
      OxfordPets    & Pet classification                     & Different breeds of cats, dogs, etc. \\
      StanfordCars  & Automobile classification              & Different car makes and models \\
      Flowers102    & Flower classification                  & Varieties of different flowers such as roses, sunflowers, etc. \\
      Food101       & Food image classification             & Pizza, burgers, sushi, etc. \\
      FGVCAircraft  & Fine-grained image classification      & Aircraft of different manufacturers and models \\
      EuroSAT       & Classification of geological features  & Different types of remote sensing images of cities, forests, etc. \\
      \bottomrule
    \end{tabularx}
    \label{table1}
  \end{table*}
The CLIP \cite{[1]} model, pre-trained with large-scale image-text pairs dataset, can effectively align image and text representations in a shared feature space, showcasing strong zero-shot learning performance across various visual tasks. Yet, the fact that its training data is mainly in English leads to a degradation of its performance when processing downstream tasks in other languages. To extend the CLIP model for languages, researchers have carried out various attempts, including customized versions of CLIP for specific languages, such as adaptations for Italian \cite{[19]} and Chinese \cite{[20],[10],[13],[21]}, augmenting the model's multilingual capability through large-scale bilingual contrastive learning and extending CLIP's multilingual capability under small scale data by employing both teacher learning and contrastive learning approaches. These works have greatly advanced CLIP's application in multilingual downstream tasks, yet they have overlooked the models’ high memory usage, which hinders their effective deployment on edge devices in practical, everyday scenarios.

Recently, with the continuous evolution and expansion of deep learning models, the research on deploying large-scale models to edge devices has increasingly become a focal point of interest. Within this domain, Knowledge Distillation (KD) \cite{[22]} has been affirmed as a promising technique for compressing large models (referred to as 'teacher models') into smaller versions (known as 'student models') that require fewer parameters and computational power while still achieving competitive results in downstream tasks. Distillation within singular modalities, such as vision or language, has been extensively explored. For example, in the field of image distillation, KDFM \cite{[23]} proposes a method to improve the effect of knowledge distillation by learning the feature mapping of a teacher network and demonstrates that the model size can be significantly reduced while maintaining the model accuracy by using generative adversarial networks and shared classifiers. This demonstrates that in image processing tasks, distillation can not only compress the model but also maintain or even improve the performance by finely shifting the feature representation. In the text domain, Xu et al. \cite{[24]} summarize how knowledge distillation makes it possible to transfer knowledge from large language models to smaller models, easing the burden of model deployment while preserving language comprehension and generation. These scholarly works provide robust pathways for deploying CLIP and its variants in other languages onto edge devices, providing a sound basis for model refinement for practical deployment in multilingual scenarios.

While considering the challenge of effectively aligning image and text features after the distillation of image and text encoders, we have incorporated insights from related work on aligning these features, adopting a contrastive learning approach to further improve the model's performance. For instance, XMC-GAN \cite{[25]} employs contrastive loss to enhance mutual information between images and texts, ensuring not only the realism of images but also their consistency with their descriptions. Moreover, UniCL \cite{[26]} extends upon this with a bidirectional learning objective, utilizing relationships within image-text-label triplets to optimize alignment across modalities. These methods are especially relevant to our goal of refining distilled features following the distillation of image and text encoders, to ensure their consistency and effective pairing. By understanding these contrastive learning strategies, we aim to address potential alignment issues encountered during distillation and improve the model's overall performance in multilingual settings.

\begin{table*}[t]
  \caption{This table presents a comparative analysis of the performance metrics for the baseline models, original CLIP, teacher model AltCLIP, and DC-CLIP method across eight English datasets. The \textbf{best} and \underline{second-best} results between our model and the baselines are prominently highlighted.}
    \centering
    \resizebox{1.0\textwidth}{!}
    {
        \begin{tabular}{lccccccccc}
          \toprule
          Method & Type & ImageNet & Caltech101 & FGVCAircraft & StanfordCars & EuroSAT & Food101 & Flowers102 & OxfordPets \\
          \toprule
          CLIP(224M) & Original & 58.19	& 80.40	& 16.95	& 53.94	& 38.16	& 80.81	& 66.02	& 85.77 \\
          AltCLIP(3.22G) & Large-size & 73.83	& 94.72	& 29.91	& 77.19	& 62.79	& 90.85	& 75.72	& 93.43 \\
          \toprule
          CNCLIP(294M) & \multirow{4}{*}{Tiny-size} & 11.038	& \underline{58.09}	& 3.75	& \underline{20.75}	& \underline{16.2}	& \underline{24.12} & \underline{13.14}	& \textbf{19.54} \\
          TaiyiCLIP(391M) & & \underline{14.84}	& 52.87	& \underline{5.94}	& \textbf{21.45}	& 13.28	& 23.71	& 7.91	& 10.38 \\
          ChineseCLIP(726M) & & 2.064	& 23.36	& 0.93	& 2.69	& 10.26	& 3.35	& 1.72	& 4.87 \\
          DC-CLIP(301M) & & \textbf{42.79}	& \textbf{73.75}	& \textbf{6.24}	& 16.54	& \textbf{64.32}	& \textbf{46.91}	& \textbf{26.43}	& \underline{17.23} \\
          \toprule
        \end{tabular}
    }
    \label{table2}
\end{table*}

\begin{table*}[t]
  \caption{This table presents a comparative analysis of the performance metrics for the baseline models, original CLIP, teacher model AltCLIP, and DC-CLIP method across eight Chinese datasets. The \textbf{best} and \underline{second-best} results between our model and the baselines are prominently highlighted.}
    \centering
    \resizebox{1.0\textwidth}{!}
    {
        \begin{tabular}{lccccccccc}
          \toprule
          Method & Type & ImageNet & Caltech101 & FGVCAircraft & StanfordCars & EuroSAT & Food101 & Flowers102 & OxfordPets \\
          \toprule
          CLIP(224M) & Original & 1.428	& 9.12	& 6.72	& 16.55	& 26.64	& 1.49	& 2.04	& 4.71 \\
          AltCLIP(3.22G) & Large-size & 56.914	& 81.04	& 29.19	& 69.56	& 55.38	& 73.91	& 49.44	& 76.94 \\
          \toprule
          CNCLIP(294M) & \multirow{4}{*}{Tiny-size} & 32.262	& \underline{74.37}	& 6.9	& \underline{27.71}	& 22.52	& 39.93	& \underline{33.91}	& \underline{54.05} \\
          TaiyiCLIP(391M) &   &\textbf{40.50}	& 71.67	& \underline{7.13}	& \textbf{34.75}	& \underline{48.4}	& \underline{42.37}	& \textbf{48.11}	& \textbf{56.25} \\
          ChineseCLIP(726M) &   & 17.394	& 52.33	& 2.91	& 11.82	& 28.64	& 19.81	& 8.60	& 10.46 \\
          DC-CLIP(301M)&  & \underline{38.142}	& \textbf{75.44}	& \textbf{7.62}	& 13.90	& \textbf{53.42}	& \textbf{43.89}	& 16.57	& 15.67 \\
          \toprule
        \end{tabular}
    }
    \label{table3}
\end{table*}

\section{Method}
In this work, we aim to ensure the model's adaptability to both English and Chinese contexts by employing the English and Chinese versions of AltCLIP as the teacher model. We introduce an innovative two-stage approach to compress the image-language representation model, making it suitable for deployment on edge devices in multilingual environments. Initially, following the methodology of Wei et al. \cite{[27]}, we utilize feature distillation for both the image and text encoders separately. This initial step is crucial for maintaining the teacher model's performance while significantly reducing the model's size and computational demands. Next, we engage in further aligning the image and text features through contrastive learning, using small-scale image-text pairs dataset. Our comprehensive training procedure is depicted in Figure 1.

\subsection{Vision-Language Feature Distillation}
\subsubsection{Distilled Image Encoder}
During the feature distillation phase for the image encoder, as shown on the left side of Figure 1a, we chose the image encoder of AltCLIP as the teacher model to enable the student model to learn robust image encoding abilities. Additionally, to reduce training time, we opted for the ResNet50 \cite{[28]} image encoder trained in CLIP as the student model. This phase leverages the teacher model's superior image-understanding capabilities to guide the student model's learning process. Diverging from traditional logic distillation, our strategy employs output feature maps from pre-trained models as distillation targets, allowing the use of any pre-trained model, even those without Logit output. To ensure the comparability of feature maps between teacher and student models, we apply the same augmented view to each raw image. Additionally, a 1 × 1 convolution layer is applied on top of the student network, allowing different dimensions of output feature maps between the teacher and student models.

To precisely gauge feature consistency between teacher and student models, we implement a smooth L1 loss function \cite{[29]}, offering enhanced robustness against outliers compared to conventional L1 loss. This function effectively mitigates potential overfitting issues in the distillation process and incorporates an adjustable hyperparameter for linear attenuation at minor errors, thereby bolstering training process stability and efficiency. Through this distillation approach, the student model assimilates a more nuanced and enriched image representation. The distillation loss is as follows:
\begin{equation}
  \begin{aligned}
  &f_s=g\left(ImageEncoder_{student}\left(I\right)\right) \\
  &f_t=ImageEncoder_{teacher}\left(I\right) \\
  &Loss_{i} = Smooth L1\left ( f_{s},f_{t} \right )
  \end{aligned}
\end{equation}
where I is the input image and g is the convolution layer of 1×1.

\subsubsection{Distilled Text Encoder}
In the distillation of text encoder stage, we distill AltCLIP's text encoder using roughly the same distillation strategy as in the distillation of image encoder stage. The difference is that here, considering that the model needs to be adapted to both Chinese and English contexts, we use the mini-model obtained from the distillation of the XLM-R \cite{[31]} model by Wang et al. \cite{[30]} as the student model. Meanwhile, to convert the output of the student model to the same output dimension as the teacher encoder, we add a fully connected layer behind the student model. As shown on the right side of Figure 1a, the distillation loss is as follows:
\begin{equation}
  \begin{aligned}
  &W_s=L\left(TextEncoder_{student}\left(T\right)\right) \\
  &W_t=TextEncoder_{teacher}\left(T\right) \\
  &Loss_{t} = Smooth L1\left ( w_{s},w_{t} \right )
  \end{aligned}
\end{equation}
where T is the input text and L is the fully connected layer.

\begin{table*}[t]
  \caption{This table presents a performance comparison between DC-CLIP and TinyCLIP trained on different training datasets, where TinyCLIP* denotes the model trained on the English image-text pairs dataset organized in this paper.}
    \centering
    \resizebox{1.0\textwidth}{!}
    {
        \begin{tabular}{lcccccccc}
          \toprule
          Method & ImageNet & Caltech101 & FGVCAircraft & StanfordCars & EuroSAT & Food101 & Flowers102 & OxfordPets \\
          \toprule
          TinyCLIP & 41.08	& 71.48	& \textbf{6.87}	& 7.01	& 22.64	& \textbf{57.05}	& \textbf{57.90}	& \textbf{45.22} \\
          TinyCLIP* & 16.32	& 56.25	& 2.46	& 1.77	& 19.70	& 11.31	& 15.63	& 40.37 \\
          \toprule
          DC-CLIP & \textbf{42.782}	& \textbf{73.75}	& 6.24	& \textbf{16.54}	& \textbf{64.32}	& 46.91	& 26.43	& 17.23 \\
          \toprule
        \end{tabular}
    }
    \label{table6}
\end{table*}
\subsection{Feature Alignment}
This training stage aims to further align image and text features through contrastive learning with Chinese and English image-text pairs dataset. As shown in Figure 1b, following the LiT \cite{[32]} approach, we freeze the distilled image encoder at training time and only update the distilled text encoder's parameters. Alignment between the output projections of the image and text encoders is achieved using a contrastive loss, known as InfoNCE loss \cite{[33]}, paralleling prior research methodologies. Specifically, given N image text pairs (x, y), where x is the image and y is the text description. Then the contrastive loss from image to text is:
\begin{equation}
  \begin{aligned}
  &\mathcal{L}_{\text{I-T}} = - \log \frac{\exp(\text{cos}(f_{\text{s}}(x), w_{\text{s}}(y)))}{\sum_{y' \in Y} \exp(\text{cos}(f_{\text{s}}(x), w_{\text{s}}(y')))}
  \end{aligned}
\end{equation}
where $f_s\left(x\right)$ is the image feature, $w_s\left(y\right)$ is the text feature, $cos\left(.,.\right)$ denotes the cosine similarity, and $Y$ is the set of text descriptions.
The contrastive loss from text to image is:
\begin{equation}
  \begin{aligned}
  &\mathcal{L}_{\text{T-I}} = - \log \frac{\exp(\text{cos}(w_{\text{s}}(y), f_{\text{s}}(x)))}{\sum_{x' \in X} \exp(\text{sim}(w_{\text{s}}(y), f_{\text{s}}(x')))}
  \end{aligned}
\end{equation}
where $f_s\left(x\right)$ is the image feature,$ w_s\left(y\right)$ is the text feature, $cos\left(.,.\right)$ denotes the cosine similarity, and $X$ is the set of images.
The total loss of comparative learning is:
\begin{equation}
  \begin{aligned}
  &\mathcal{Loss}_{\text{CL}} =\frac{1}{2} \left ( \mathcal{L}_{\text{T-I}} + \mathcal{L}_{\text{T-I}} \right )
  \end{aligned}
\end{equation}
By minimizing the loss of contrastive learning, the features between images and text are further aligned, thus improving the performance of the model.

\section{Experiment}
\subsection{Experiment Setting}
\subsubsection{Training Datasets}
During the distillation of image encoder stage, we organized the training sets of Imagenet-1k \cite{[34]}, Caltech101 \cite{[35]}, FGVC Aircraft \cite{[36]}, Stanford Cars \cite{[37]}, EuroSAT \cite{[38]}, Food-101 \cite{[39]}, UCF101 \cite{[40]}, and Oxford Flowers 102 \cite{[41]} as the dataset for training the image encoder. The distillation text encoder stage utilizes the translation2019zh Chinese-English translation dataset, which consists of approximately 5.2 million Chinese-English parallel corpora. The training set contains 5.16 million corpora, while the validation set contains 39,000 corpora. Both the training set and validation set are formatted in JSON. In the comparative learning stage, we organized 1.4 million Chinese image-text pairs from CLIP-Chinese and 1.5 million English image-text pairs from LAION 5B. The English image-text pairs dataset from LAION 5B \cite{[42]} is a randomly selected subset of data that has been filtered by the Laion Aesthetic v2 model with a threshold score exceeding 6.25 producing a small-scale, high-quality dataset of Chinese-English image-text pairs.

\begin{table*}[t]
  \caption{This table shows the comparative analysis of the performance metrics of the DC-CLIP method before and after the comparative learning on the eight datasets, with the best performing results highlighted.}
    \centering
    \resizebox{1.0\textwidth}{!}
    {
        \begin{tabular}{lccccccccc}
          \toprule
          Method  & language & ImageNet & Caltech101 & FGVCAircraft & StanfordCars & EuroSAT & Food101 & Flowers102 & OxfordPets \\
          \toprule
          DC-CLIP-P& \multirow{2}{*}{English}& 41.912 & 71.87 & 5.55 & 15.69 & 62.75 & 45.56 & 26.12 & 12.7  \\
          DC-CLIP  &     & \textbf{42.782} & \textbf{73.75} & \textbf{6.24} & \textbf{16.54} & \textbf{64.32} & \textbf{46.91} & \textbf{26.43} & \textbf{17.23} \\
          \toprule
          DC-CLIP-P & \multirow{2}{*}{Chinese} & 37.666 & 73.59 & 6.24 & 10.81 & 49.76 & 43.57 & 16.43 & 11.04 \\
          DC-CLIP   &          & \textbf{38.142} & \textbf{75.44} & \textbf{7.62} & \textbf{13.9} & \textbf{53.42} & \textbf{43.89} & \textbf{16.57} & \textbf{15.67} \\
          \toprule
        \end{tabular}
    }
    \label{table4}
\end{table*}

\begin{table*}[t]
  \caption{This table shows the results of the first stage distillation image encoder  after distillation using different image encoders, with the best-performing result highlighted.}
    \centering
    \resizebox{1.0\textwidth}{!}
    {
        \begin{tabular}{lcccccccc}
          \toprule
          Method & ImageNet & Caltech101 & FGVCAircraft & StanfordCars & EuroSAT & Food101 & Flowers102 & OxfordPets\\
          \toprule
          ViT & 40.438	& 55.375	& 1.41	& 2.08	& 22.08	& 0.59	& 6.04 & 66.45\\
          ResNet50 & \textbf{67.324}	& \textbf{82.15}	& \textbf{8.04}	& \textbf{15.96}	& \textbf{19.56}	& \textbf{42.17}	& \textbf{16.20} & \textbf{71.42} \\
          \toprule
        \end{tabular}
    }
    \label{table5}
\end{table*}
\subsubsection{Test Datasets}
Our DC-CLIP was tested on eight public image classification datasets to showcase its effectiveness in zero-shot classification experiments. The datasets consist of two generic-objects datasets, ImageNet and Caltech101; five fine-grained classification datasets, OxfordPets \cite{[43]}, StanfordCars, Flowers102, Food101, and FGVCAircraft; and a satellite image dataset, EuroSAT. Details of the datasets are listed in Table 1. To evaluate the Chinese CLIP on the dataset, we first transform the datasets for Chinese models by translating labels and prompts into Chinese. Specifically, we translate the text descriptions of the labels and the templates of the manual prompts into Chinese. The labels in caltech101, such as "car" and "dog," are translated into Chinese manually. Special cases, like the labels in FGVC-Aircraft, can be challenging to translate or transliterate. We research the names on Google to discover the optimal Chinese name for each label. However, we cannot guarantee that we have the best Chinese translation. Additionally, Chinese pre-trained models may struggle to comprehend certain concepts, resulting in subpar performance in related tasks. We utilize template translations from the ELEVATER \cite{[46]} toolkit for certain datasets and from the OpenAI CLIP templates for other datasets.


\subsubsection{Baselines}
To evaluate our approach, we compared DC-CLIP against the original CLIP, AltCLIP, and other variants of comparable parameter scale on both Chinese and English datasets. Specifically, we analyzed three CLIP variants adapted to different linguistic contexts: Taiyi-CLIP, CN-CLIP, and ChineseCLIP. Taiyi-CLIP employs Chinese-roberta-wwm \cite{[44]} as its text encoder and ViT-B/32 \cite{[45]} from CLIP as its image encoder. CN-CLIP utilizes the RBT3 architecture for its text encoder and the ResNet50 architecture for its image encoder. ChineseCLIP integrates a Bert architecture for text encoding, pairing it with ViT-B/32 from CLIP for image encoding. The memory footprint of DC-CLIP and these variant CLIP models is approximately 500M. Meanwhile, to verify that DC-CLIP achieves good performance with only small-scale data, we compared the model replicated on our small-scale English image-text pairs dataset, referred to as TinyCLIP*, with the original TinyCLIP trained on a large-scale dataset. This comparison was conducted on the English datasets.

\subsubsection{Parameter Settings}
In the stage of distilling the image encoder, we adopted the hyperparameter settings as outlined by Wei et al., setting the image size to 224, batch size to 128, epochs to 30, learning rate to 3e-4, and weight decay to 0.05. For the text encoder distillation stage, the batch size was set to 256, with epochs set to 10 and a learning rate of 8e-5.  In the contrastive learning phase, we set a batch size of 256, processing the data once, with a learning rate of 2e-6. All models were trained using two NVIDIA RTX4090 GPUs.

\subsection{Performance Evaluation}
The results obtained by different methods on 8 English datasets and 8 Chinese datasets are listed in Table 2 and Table 3. From the results, compared with the baselines, DC-CLIP significantly outperforms the baseline model in accuracy across the majority of datasets in the English context (in 6 out of 8 cases), with only slight underperformance noted in the StanfordCars and OxfordPets datasets. But the performance is only slightly degraded, which may be caused by the lack of text about the StanfordCars and OxfordPets fine-grained categorical data in the English text data when distilling the text encoder. In the Chinese context, DC-CLIP demonstrates superior accuracy over CN-CLIP on five datasets (ImageNet, Caltech101, FGVC, EuroSAT, Food101), exceeds Taiyi-CLIP in four datasets (Caltech101, FGVCAircraft, EuroSAT, Food101), and comprehensively outperforms ChineseCLIP across all assessed datasets. A possible reason for the weak performance on the other evaluated datasets is that the Chinese dataset in the distillation text encoder stage is of poorer quality and contains less finely categorized text. In conclusion, the results show that our model holds competitive advantages over baseline models in the Chinese context and exhibits superior performance in the English context, particularly on the generic-objects dataset of ImageNet-1k. In conclusion, these results validate the effectiveness of our idea.

Furthermore, to validate the advantages of our approach in achieving competitive performance with only small-scale dataset and the effectiveness of the two-stage compression model strategy, we replicated the TinyCLIP \cite{[47]} model, referred to as TinyCLIP*, on the English image-text pairs dataset we organized. We then compared its performance across eight English evaluation datasets with the original TinyCLIP model trained on a dataset of over a billion samples. As shown in Table 4, compared to TinyCLIP*, DC-CLIP exhibits lower performance only on the OxfordPets dataset, outperforming TinyCLIP* on the other datasets. In comparison to TinyCLIP, DC-CLIP demonstrates performance advantages on four datasets (ImageNet, Caltech101, StanfordCars, EuroSAT), showcasing a certain level of competitiveness. In conclusion, DC-CLIP holds the advantage of achieving good performance with only a small-scale dataset and further validates the effectiveness of the two-stage compression model strategy.

\subsection{Ablation Studies}
\subsubsection{Impact of contrastive learning}
To assess the efficacy of the contrastive learning, we analyzed the performance of the DC-CLIP model both with and without the contrastive learning on 8 different test datasets. Table 5 illustrates that DC-CLIP performs better than DC-CLIP prior to contrastive learning on all test sets. This demonstrates the effectiveness of contrastive learning, which aligns image and text features, improves the model's ability to generalize across datasets, mitigates data bias, and enhances the model's robustness and generalization performance. At the same time, it also proves the importance of the contrastive learning strategy adopted by DC-CLIP.
\subsubsection{Impact of different image encoder}
In the stage of distilling the image encoder, in order to explore the effect of choosing different image encoders as the student model on the distillation effect, we chose the vit architecture with 6-layer transformer and Resnet50 as the image encoder respectively. The distillation results are shown in Table 6, which shows that under the same size of training data, the selection of Resnet50 as the student model is more effective. This indicates that Resnet50 can learn the effective knowledge of the teacher model better under a relatively small training dataset. Meanwhile, in order to enable the model to converge faster, we choose the pre-trained image encoder ResNet50 of CLIP as the initialized student model.

\section{Conclusion}
In this study, we developed and evaluated a comprehensive model distillation approach, DC-CLIP, that emphasizes bilingual adaptability, innovatively uses output feature mapping as a distillation target, and combines contrastive learning to improve model performance on a limited training dataset. Our findings reveal that this methodology not only facilitates the adaptation of models to both English and Chinese contexts but also significantly enriches the knowledge transfer process between teacher and student models. By focusing on output feature maps rather than traditional logits or simplified feature vectors, we were able to capture a more nuanced representation of information. The inclusion of contrastive learning further augmented this process, enabling a more precise alignment of features across different modalities and languages.

However, there are some limitations to our approach. While our DC-CLIP achieves state-of-the-art performance in the English environment, it only attains competitive performance compared to the baseline model in the Chinese environment. Therefore, our future work could delve deeper into optimizing feature mapping distillation techniques and assess the suitability of our approach in more resource-constrained computing environments. Additionally, exploring the extension of this approach to other languages and domains is warranted. The potential for enhancing model understanding and facilitating interaction across diverse cultures and languages is vast, and our work serves as a foundational step for further explorations in this direction.

\bibliography{Literatures}

\end{document}